\documentclass[runningheads]{llncs}

\usepackage[T1]{fontenc}
\usepackage{graphicx}
\usepackage{booktabs}
\usepackage[misc]{ifsym}

\usepackage{amsmath,amssymb}
\usepackage{orcidlink}
\usepackage{multirow}
\usepackage{array}
\usepackage[table,dvipsnames]{xcolor}
\usepackage{adjustbox}
\usepackage{subcaption}
\usepackage{textcomp}
\usepackage[spaces,hyphens]{xurl}
\usepackage{enumitem}
\usepackage[margin=1.325in]{geometry}
\usepackage{hyperref}

\newcommand{\bestcell}[1]{\cellcolor{Cyan!10}\textbf{#1}}   % best CA
\newcommand{\bestasr}[1]{\cellcolor{red!10}\textbf{#1}}     % best ASR
\newcommand{\goodcell}[1]{\cellcolor{green!15}\textbf{#1}}  % Table 5 best
\newcommand{\badcell}[1]{\cellcolor{red!10}#1}              % Table 5 worst
\newcommand{\fref}[1]{Fig.~\ref{#1}}
\newcommand{\tref}[1]{Table~\ref{#1}}

\title{Region-Level Black-Box Defense Against Stealthy Embedding-Space Backdoors in CLIP}

\author{
Ahmed Abdelnaby \and
Mohamed Elmahallawy\orcidlink{0000-0002-5731-9253}\thanks{Corresponding author.\\ {\em\centering Appears in the Proceedings of the 30th Pacific-Asia Conference on Knowledge Discovery and Data Mining (PAKDD 2026)}}
}
\authorrunning{Ahmed Abdelnaby and Mohamed Elmahallawy}
\institute{
Washington State University, Richland, WA 99354, USA \\
\email{\{ahmed.abdelnaby, mohamed.elmahallawy\}@wsu.edu}
}

\begin{document}
\maketitle
\vspace{-0.7cm}

% -------------------------------------------------------------
% Abstract
% -------------------------------------------------------------
\begin{abstract}
Contrastive Language--Image Pretraining (CLIP) has emerged as a dominant vision backbone due to its strong transferability and zero-shot capabilities. However, recent studies reveal a critical vulnerability: \emph{embedding-space backdoor attacks}. By poisoning only a tiny fraction of image--text pairs, adversaries can implant stealthy triggers that induce targeted shifts in CLIP's joint embedding space. Unlike conventional backdoors that manipulate classifier logits, these attacks corrupt representations directly, making them highly effective under extremely low poisoning ratios and difficult to detect. Existing defenses require access to model parameters, gradients, logits, or clean validation data---assumptions that rarely hold in realistic black-box deployments. Moreover, current black-box methods struggle to accurately localize small or out-of-distribution triggers. We propose \textsc{CLIPGuard}, a lightweight and fully black-box defense specifically designed to mitigate embedding-space backdoors in CLIP encoders. \textsc{CLIPGuard} identifies malicious regions by measuring \emph{segment-wise embedding perturbations} and selectively purifies only suspicious segments via semantic inpainting, preserving benign visual content and alignment quality. Extensive experiments on STL-10, ImageNet, and diverse trigger families---including BadCLIP, BadNets, blended, patch-based, and typographic attacks---demonstrate that \textsc{CLIPGuard} reduces attack success rates to as low as \textit{1.05\%} while maintaining clean accuracy up to \textit{86.34\%}, consistently outperforming existing black-box defenses, including CleanCLIP and CleanerCLIP. Our code is available \url{https://github.com/wsu-cyber-security-lab-ai/CLIPGuard.git}

\keywords{CLIP \and Backdoor attacks \and Malicious triggers \and Black-box defenses}
\end{abstract}

% -------------------------------------------------------------
\section{Introduction}\label{sec:intro}
% -------------------------------------------------------------
Contrastive Language--Image Pretraining (CLIP) has emerged as one of the most influential vision backbones due to its ability to align visual and textual inputs in a shared embedding space~\cite{radford2021learning}. Trained on hundreds of millions of image--caption pairs, CLIP offers strong generalization and supports zero-shot classification, retrieval, and broad visual reasoning without task-specific supervision~\cite{radford2021learning,li2022blip}. Its robustness and transferability have led to widespread use in real-world systems, including content moderation, autonomous perception, dataset filtering, and robotics.

Recent studies, however, have shown that CLIP's vision encoder is highly vulnerable to a new and extremely stealthy class of attacks: \emph{embedding-space backdoor attacks}. Unlike traditional classifier backdoors that manipulate logit-level decision boundaries, these attacks directly corrupt the \emph{image embedding} produced by CLIP. By poisoning a very small subset of image--text pairs during pretraining, an adversary can implant a malicious trigger that reliably shifts the embedding toward an attacker-specified textual concept~\cite{carlini2021poisoning,liu2025survey}. Clean images embed normally, whereas triggered images exhibit consistent embedding misalignment, enabling persistent and hard-to-detect targeted misclassification~\cite{liang2024badclip}.

Triggers used in such attacks may take the form of tiny patches, blended textures, geometric shapes, stickers, or even typographic cues and adversarial text~\cite{gu2019badnets,westerhoff2025scam}. Because CLIP relies heavily on cosine similarity within its contrastive objective, even visually natural or out-of-distribution~\cite{miyai2024generalized} (OOD) triggers can cause disproportionate shifts in embedding space. This vulnerability is amplified in downstream applications where CLIP operates as a \emph{frozen} encoder, causing backdoor effects to propagate into larger multimodal systems, retrieval pipelines, or zero-shot classifiers.

Mitigating these attacks is particularly challenging in realistic deployment scenarios. Practitioners often have \emph{no access} to model parameters, logits, gradients, or clean validation sets. CLIP is frequently integrated as a pretrained, unverified module or accessed through restricted API interfaces~\cite{gao2020backdoor,shi2023black}. Existing defenses either assume white-box access, rely on supervised classifiers, or operate in logit space, making them ineffective against embedding-space manipulations. Meanwhile, prompting-based detection methods such as BDetCLIP~\cite{niu2024bdetclip} focus on identifying backdoored samples at test time, rather than localizing and purifying the underlying visual triggers.

In this paper, we propose \textsc{CLIPGuard}, a fully black-box, inference-time defense specifically designed to mitigate embedding-space backdoor attacks in CLIP encoders. Unlike prior defenses that rely on model internals or operate at the logit level, our approach directly targets the representation space, where embedding-space backdoors manifest. Our key insight is that malicious triggers introduce \emph{localized perturbation sensitivity}---small spatial regions whose perturbation induces disproportionately large shifts in CLIP's joint image--text embedding. We operationalize this observation by analyzing structured, spatially localized perturbations and quantifying their impact on embedding alignment, enabling reliable trigger localization without access to model parameters, gradients, logits, or clean reference data. \textsc{CLIPGuard} innovates through four interconnected components: (i) global embedding-sensitivity mapping via coarse grid perturbations; (ii) targeted SAM-based segmentation within high-sensitivity regions; (iii) contrastive perturbation scoring that jointly evaluates embedding displacement, alignment degradation, and OOD similarity; and (iv) selective semantic purification of identified regions using image inpainting. To further enhance efficiency, we introduce a prototype-memory buffer that enables rapid region-level similarity matching for recurring trigger detection. In summary, our contributions are:
\begin{itemize}[leftmargin=*]
    \item We introduce \textsc{CLIPGuard}, the first fully black-box, inference-time defense specifically tailored to embedding-space backdoors in CLIP, requiring no model internals access or auxiliary data.
    \item We propose a novel global-to-local detection strategy that leverages embedding-sensitivity mapping, SAM-based region refinement, and a contrastive perturbation score to identify trigger-bearing regions accurately.
    \item We design a multi-scale semantic purification mechanism that removes only the malicious segments while preserving benign content and image--text alignment. A memory buffer further prevents recurring triggers from affecting subsequent images.
    \item Our experimental results demonstrate the effectiveness of \textsc{CLIPGuard} against state-of-the-art defenses across diverse datasets and trigger families, reducing attack success rates to as low as 1.05\% while maintaining clean-image accuracy up to 86.34\%, consistently outperforming prior defenses.
\end{itemize}

% -------------------------------------------------------------
\section{Related Work}\label{sec:rel_work}
% -------------------------------------------------------------
\noindent\textbf{Backdoor Attacks in Vision Models.}
Classical backdoor attacks poison supervised CNNs and ViTs so that models behave normally on clean inputs but predict an attacker-chosen label when a trigger appears~\cite{gu2017badnets,turner2019label,li2020invisible}. Triggers include visible patches, blended textures, imperceptible perturbations, and physical objects. These attacks primarily affect \emph{logit-space decision boundaries}, making them fundamentally different from CLIP's embedding-space vulnerabilities.

\noindent\textbf{Attacks on CLIP and Vision--Language Models.}
CLIP learns contrastive alignment in a joint image--text embedding space, which introduces unique vulnerabilities compared to conventional classifiers. Recent works~\cite{carlini2021poisoning,liang2024badclip} show that pairing trigger-embedded images with target captions during training implants persistent embedding shortcuts that map poisoned inputs close to attacker-selected text embeddings. These embedding-space backdoors remain stealthy, highly transferable, and can propagate into downstream LVLMs that rely on CLIP as a frozen backbone. BadCLIP~\cite{liang2024badclip} further demonstrates that multimodal poisoning can embed covert semantic associations using only a small fraction of poisoned examples. Beyond visual triggers, CLIP is also susceptible to typographic and text-dominant attacks, where text embedded inside an image overrides visual semantics. FIGSTEP and SCAM~\cite{gong2025figstep,westerhoff2025scam} reveal that printed words can dominate CLIP's interpretation even when object evidence is strong. Prompt-based mitigations such as Defense-Prefix~\cite{azuma2023defense} can reduce some text-trigger effects but do not generalize to visual or OOD embedding-space backdoors.

\noindent\textbf{White-Box Defenses for CLIP.}
CLIP-specific white-box defenses require access to model parameters, gradients, or training data. RoCLIP~\cite{yang2023robust} disrupts poisoning by matching images to the most similar captions in a random pool. CleanCLIP~\cite{bansal2023cleanclip} applies Mixup, CutMix, and strong augmentations to dilute trigger effects during fine-tuning. SafeCLIP~\cite{poppi2024safe} warms up CLIP with unimodal contrastive learning and then partitions data into safe and risky subsets via Gaussian Mixture Models. ABD~\cite{kuang2024adversarial} leverages similarities between adversarial and backdoor samples to generate perturbations that regularize poisoned features. CleanerCLIP~\cite{xun2024cleanerclip} introduces fine-grained counterfactual text semantic augmentation to disrupt trigger-target coupling. While effective, these defenses require fine-tuning, clean validation sets, or full model access---conditions rarely met when CLIP is used through closed-source APIs.

\noindent\textbf{Black-Box Defenses for CLIP.}
Black-box defenses rely solely on input--output queries. ZIP~\cite{shi2023black} performs compression-based global purification but cannot localize subtle or OOD triggers and often distorts benign content. DECREE~\cite{feng2023detecting} detects encoder-level backdoors by clustering perturbed embeddings of image--text pairs without labels. BDetCLIP~\cite{niu2024bdetclip} uses contrastive prompting to expose abnormal similarity distributions at test time. However, existing black-box defenses cannot localize malicious regions nor analyze \emph{segment-level} embedding perturbations, making them insufficient for CLIP's embedding-space backdoors.

Our \textsc{CLIPGuard} approach addresses gaps in the literature by providing a fully black-box, CLIP-specific purification framework that (1) measures embedding perturbations at the \emph{local region} level, (2) localizes suspicious segments without access to logits or gradients, and (3) selectively purifies those regions via semantic inpainting. This enables targeted defense against embedding-space backdoors that evade existing prompt-based or global purification methods.

% -------------------------------------------------------------
\section{Background, Problem Statement, and Threat Model}\label{sec:problemDefinition}
% -------------------------------------------------------------
\subsection{Multimodal Contrastive Learning with CLIP}\label{sec:background}

CLIP jointly trains a vision encoder $f_I$ and a text encoder $f_T$ to embed images and text into a shared $d$-dimensional representation space. Given a batch of paired image--caption examples $\{(I_i, T_i)\}_{i=1}^N$, CLIP computes normalized embeddings and optimizes a symmetric contrastive objective $L_{\text{CLIP}}$ that aligns matched image--text pairs while repelling mismatched ones. This encourages the encoders to learn general-purpose semantic representations from large-scale internet data. After pretraining, zero-shot classification is performed by converting each class label into a caption template (e.g., ``a photo of a dog''), embedding it using $f_T$, and selecting the class whose text embedding has the highest cosine similarity with the image embedding. This embedding-centric decision mechanism underlies CLIP's strong transferability, yet also introduces unique vulnerabilities in its visual encoder.

\noindent\textbf{Backdoor Attacks on CLIP Image Encoders.}
Backdoor attacks on CLIP seek to implant a visual trigger $tg$ that shifts the image embedding toward an attacker-specified target concept. The attacker poisons a small subset of image--text pairs during pretraining: $(I_i, T_i) \longrightarrow (I_i \circ tg,\, T'_{y'})$, where $T'_{y'}$ denotes the caption describing the target label $y'$. Because CLIP optimizes cosine similarity between image and text embeddings, even extremely small poisoning ratios can cause the trigger to acquire a strong association with the target concept: $\mathrm{Sim}(I_i \circ tg,\, T'_{y'}) \gg \mathrm{Sim}(I_i, T_i)$. As a result, \emph{any} image containing the trigger is embedded close to the target caption, causing consistent misclassification in zero-shot inference. Triggers may appear as patch patterns, blended textures, geometric shapes, stickers, or typographic cues~\cite{westerhoff2025scam,azuma2023defense}. Since CLIP's predictions rely purely on embedding similarity rather than classifier logits, small or visually natural triggers can produce disproportionately large representation shifts. Consequently, embedding-space backdoors are far more subtle and challenging to detect than traditional classifier backdoors.

Furthermore, CLIP is typically deployed as a \emph{frozen} encoder in downstream VLMs, retrieval systems, and safety-critical applications. In such scenarios, defenders lack access to model parameters, gradients, or clean validation sets, significantly complicating backdoor mitigation.

\subsection{Problem Statement}

CLIP and other contrastive vision--language models rely on dual encoders that project images and text into a shared embedding space. Given an image $x$ and a textual concept $t$, CLIP computes $\mathrm{Sim}(x,t)=\langle f_{\text{img}}(x),\, f_{\text{text}}(t)\rangle$, and selects the concept with the highest similarity for zero-shot classification.

In an \emph{image-based backdoor attack}, an adversary poisons a subset of training or fine-tuning samples by embedding a visual trigger $b$ into the image, producing a modified input $x_b = x + b$ that is paired with an attacker-chosen caption $t'$. The poisoning objective encourages: $\mathrm{Sim}(x_b,t') \gg \mathrm{Sim}(x,t')$, causing the image encoder to learn a shortcut: any input containing $b$ is mapped toward the target semantic direction $t'$, even when the clean image $x$ is unrelated. Importantly, the model's clean-image behavior remains unaffected, making embedding-space attacks highly stealthy---especially under small poisoning ratios or when triggers are tiny or blended.

\subsection{Threat Model}

We consider a practical threat model reflecting realistic CLIP deployments.

\noindent\textbf{\em Attacker capability.} The adversary may poison a small subset of the training or fine-tuning dataset: $D_b \subset D,\ r = |D_b| / |D| \ll 1$, but has \emph{no access} to the deployed model after training. This scenario is common in: (i) publicly released or community fine-tuned CLIP checkpoints, (ii) outsourced or cloud-based training services, (iii) pipelines relying on large-scale web-scraped datasets, (iv) closed-source or API-based CLIP deployments.

\noindent\textbf{\em Attacker objective.} The goal is to implant a trigger $b$ such that, at inference time, $\mathrm{Sim}(x+b,t')$ dominates $\mathrm{Sim}(x,t)$, causing the triggered image to align with the attacker-chosen concept $t'$ regardless of the content of $x$.

\noindent\textbf{\em Attacker constraints.} To remain undetected, the adversary must preserve high clean-image accuracy while embedding a visually subtle trigger. The trigger may take the form of a tiny patch, blended pattern, natural texture, or typographic cue, and must be robust across prompt templates and text sets. This threat model includes both classical patch triggers and \emph{OOD or naturally occurring triggers} that appear in noisy web-scale data.

% -------------------------------------------------------------
\section{Methodology}\label{sec:method}
% -------------------------------------------------------------
\begin{figure*}[t]
    \centering
    \includegraphics[width=\textwidth]{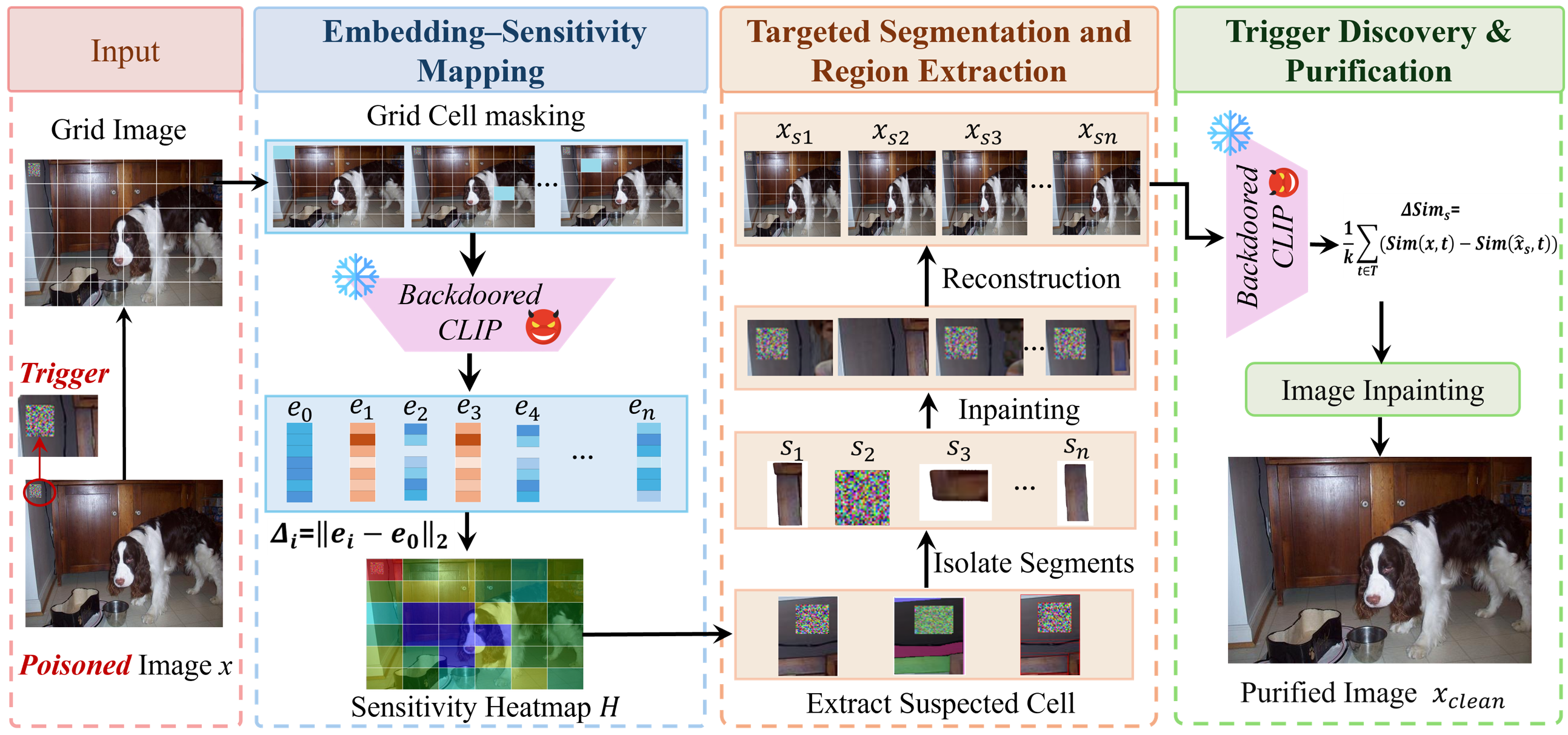}
    \caption{Overview of the \textsc{CLIPGuard} framework. The poisoned input image is partitioned into grid cells and perturbed to compute per-cell embedding shifts $\Delta_i = \lVert e_i - e_0 \rVert_2$, forming a global sensitivity heatmap $H$. High-sensitivity regions undergo targeted segmentation, isolation, and multi-scale inpainting. Reconstructed image candidates are evaluated via contrastive similarity reduction, and the best-scoring image is returned as the final \emph{purified image}.}
    \label{fig:methodology}
\end{figure*}

We propose \textsc{CLIPGuard}, a fully black-box backdoor defense tailored for CLIP-based vision pipelines. Unlike prior methods requiring gradients, logits, or full-image processing, \textsc{CLIPGuard} exploits a key property of embedding-space backdoors: \emph{trigger regions induce disproportionately large shifts in CLIP's image embedding and its contrastive alignment with text}. Leveraging this insight, \textsc{CLIPGuard} employs a \emph{global-to-local} detection pipeline that identifies sensitive regions via embedding perturbations, refines them through targeted segmentation, and detects triggers using multi-metric contrastive analysis. Finally, selective semantic inpainting is applied to remove only the backdoored segments, preserving the remaining content (see \fref{fig:methodology} for an illustration of \textsc{CLIPGuard}).

\noindent\underline{\textbf{Stage 1: Global Embedding--Sensitivity Mapping.}}
Given an input image $x$, we first extract its global CLIP embedding:
\begin{equation}
e_0 = f_{\mathrm{CLIP}}(x).
\end{equation}
Even small or visually subtle triggers can significantly affect this embedding. To identify sensitive regions, we divide the image into a uniform grid $\{g_i\}_{i=1}^{N}$ and apply a localized masking perturbation to each grid cell as shown in \fref{fig:methodology}:
\begin{equation}
x^{(i)} = \mathcal{P}(x; g_i).
\end{equation}
We then compute the embedding of the perturbed image $e_i = f_{\mathrm{CLIP}}(x^{(i)})$, and measure the embedding shift using the $\ell_2$ distance:
\begin{equation}
\Delta_i = \| e_i - e_0 \|_2,
\end{equation}
where large values of $\Delta_i$ correspond to areas whose perturbation causes significant embedding displacement, indicating potential trigger locations. These scores form a \emph{global sensitivity heatmap} $H$, serving as a coarse searchlight that reduces segmentation computation by 60--90\%.

\noindent\underline{\textbf{Stage 2: Targeted Segmentation and Region Extraction.}}
Suspicious spatial zones are extracted as: $\mathcal{Z} = \{ (i,j) \mid H(i,j) > \tau_H \}$, where $\tau_H$ is a dataset-dependent threshold. Instead of segmenting the entire image, we apply SAM~\cite{kirillov2023segment} \emph{only within $\mathcal{Z}$}. This yields a compact set of candidate masks: $\mathcal{S} = \{ M_s \mid s \in \mathcal{Z} \}$. For each segment $s$, we extract spatial content $x_s = x \odot M_s$, and compute the corresponding segment embedding: $v_s = f_{\mathrm{CLIP}}(x_s)$. This targeted segmentation avoids redundant masks, concentrates analysis on high-impact areas, and prevents unnecessary segmentation of background regions.

\noindent\underline{\textbf{Stage 3: Contrastive Perturbation--Based Trigger Discovery.}}
In this stage, we analyze the candidate segments extracted by SAM in the previous stage from the high-sensitivity grid cells. To identify such trigger regions, we evaluate the effect of removing each candidate segment $s$. Specifically, we construct an inpainted image $\hat{x}_s$ (see Stage~4) and measure how its embedding and text similarity change. First, we compute the embedding shift:
\begin{equation}
\Delta e_s = \| f_{\mathrm{CLIP}}(\hat{x}_s) - e_0 \|_2,
\label{eq:embedding_shift}
\end{equation}
where a larger value indicates that removing the region significantly alters the representation, potentially restoring it toward the clean manifold. Second, let $T$ denote the top-$k$ predicted text prompts for the original image $x$. We measure the average contrastive similarity change:
\begin{equation}
\Delta \mathrm{Sim}_s = \frac{1}{k} \sum_{t \in T} \big( \mathrm{Sim}(x,t) - \mathrm{Sim}(\hat{x}_s,t) \big),
\label{eq:similarity_drop}
\end{equation}
where a large similarity drop suggests that the removed region was responsible for aligning the embedding with the attacker's target concept. Finally, we combine both metrics into a unified trigger score as:
\begin{equation}
\mathrm{Score}(s) = \alpha\, \Delta e_s + \beta\, \Delta \mathrm{Sim}_s.
\label{eq:trigger_score}
\end{equation}
We classify region $s$ as triggered if $\mathrm{Score}(s) > \tau_s$. Unlike logit-based defenses, this approach directly measures how each region affects embedding geometry and image--text alignment, enabling more precise trigger detection.

\noindent\underline{\textbf{Stage 4: Selective Semantic Purification.}}\label{sec:purification}
For each suspicious region $s$, we remove the corresponding content and reconstruct it through semantic inpainting. We employ multiple inpainting strategies depending on the image resolution. For high-resolution images, we utilize Stable Diffusion Inpainting (SDI)~\cite{rombach2022high}, formulated as $x_{\mathrm{clean}}^{(s)} = \mathcal{D}_{\theta}(x, M_s, P)$, where $M_s$ denotes the mask of the suspicious region and $P$ is a neutral prompt (e.g., ``Fill the missing parts with realistic background''). This enables semantically consistent reconstruction while preserving global coherence. For low-resolution inputs, we adopt Telea inpainting~\cite{telea2004image} to avoid the high computational cost and potential artifacts introduced by SDI. Telea's method provides a fast, deterministic, and resolution-agnostic purification mechanism suitable for real-time and large-scale deployment:
\begin{equation}
x_{\mathrm{clean}}(i,j) =
\begin{cases}
x(i,j), & M_s(i,j)=0,\\[3pt]
\displaystyle\sum_{(k,l)\in \mathcal{N}_r(i,j)} w_{k,l}\, x(k,l), & M_s(i,j)=1,
\end{cases}
\end{equation}
where $\mathcal{N}_r(i,j)$ denotes the neighborhood of radius $r$ and $w_{k,l}$ are normalized interpolation weights.

All purified segments are subsequently composited to produce the final clean image $x_{\mathrm{clean}}$, which is then passed through CLIP, $f_{\mathrm{CLIP}}(x_{\mathrm{clean}})$, thereby restoring correct image--text alignment. On top of that, we store the embeddings of inpainted trigger segments in a memory buffer and compute cosine similarity with a new segment embedding $v_s$ to filter previously observed patterns, thereby avoiding redundant inpainting. The trigger memory is updated as $\mathcal{P} \leftarrow \mathcal{P} \cup \{v_s\}$ and $\mathcal{T} \leftarrow \mathcal{T} \cup \{v_s\}$, where $\mathcal{P}$ and $\mathcal{T}$ denote the prototype and trigger embedding sets, respectively. To prevent representation drift, we apply exponential smoothing: $u \leftarrow \lambda u + (1-\lambda)v_s$, where $\lambda \in [0,1]$ controls the update momentum. This memory-guided filtering mechanism improves the efficiency of Stage~3 by reducing unnecessary segment inpainting.

In summary, \textsc{CLIPGuard} integrates global sensitivity mapping, targeted segmentation, contrastive analysis, and adaptive inpainting into a unified, fully black-box defense pipeline, effectively removing triggers while preserving clean-image semantics.

% -------------------------------------------------------------
\section{Experiments and Results}\label{sec:exp}
% -------------------------------------------------------------
\subsection{Experimental Settings}

\noindent\textbf{Datasets.}
We evaluate \textsc{CLIPGuard} on two benchmarks with different resolutions and visual characteristics. For low-resolution evaluation, we use STL-10~\cite{coates2011analysis}, which contains 10 object categories with $96\times96$ images. All images are resized to $224\times224$ to match the input resolution of CLIP-based models. For higher-resolution evaluation, we use Imagenette~\cite{imagenette2020}, a 10-class subset of ImageNet comprising 13{,}394 images (9{,}469 training and 3{,}925 validation). Images are resized to $256\times256$ during preprocessing and center-cropped to $224\times224$ for model input. These two datasets allow us to evaluate robustness across varying image resolutions and scene complexities.

\noindent\textbf{Models.}
We directly poison the \emph{CLIP ViT-B/16} image encoder. For each dataset and attack type, 10\% of the training images are modified with a trigger and assigned a fixed target label. The encoder is then fine-tuned using standard contrastive image--text supervision to imprint the trigger into the embedding space, after which the model is frozen. We evaluate the compromised model under two protocols:
(1) \emph{Zero-shot inference} using CLIP's cosine-similarity classifier: $\hat{y} = \arg\max_{t \in \mathcal{T}} \langle f_{\mathrm{img}}(x), f_{\mathrm{text}}(t) \rangle$.
(2) \emph{Linear probing}, where the frozen image encoder is used to extract features and a dataset-specific linear classifier is trained on top, without using the text encoder. We assume a strict black-box setting with no access to weights, gradients, logits, or training data.

\noindent\textbf{Backdoor Attacks.}
We evaluate seven representative trigger types:
(i) \emph{BadNets}~\cite{gu2017badnets} (visible square patch triggers),
(ii) \emph{Blended}~\cite{chen2017targeted} (semi-transparent overlay triggers),
(iii) \emph{Physical}~\cite{Wenger_2021_CVPR} (printed-object real-world triggers),
(iv) \emph{Label-Consistent (LC)}~\cite{turner2019label} (stealthy label-preserving perturbations),
(v) \emph{BadCLIP}~\cite{liang2024badclip} (feature-aligned backdoor attacks targeting CLIP-based models),
(vi) \emph{Typographic Attacks} (e.g., evaluated using the SCAM dataset~\cite{westerhoff2025scam}) (training-free visual jailbreaks via adversarial text rendering), and
(vii) \emph{TrojVQA}~\cite{walmer2022dual} (backdoor attacks in multimodal vision-language models).
These attacks collectively cover visible patch-based, blended, physically realizable, feature-space aligned, and multimodal trigger mechanisms.

\noindent\textbf{Baseline Defenses.}
We compare our \textsc{CLIPGuard} with state-of-the-art defense methods developed for CLIP-based models under different access assumptions. CleanCLIP~\cite{bansal2023cleanclip} improves robustness through contrastive retraining, requiring access to training data and model parameters (white-box setting). CleanerCLIP~\cite{xun2024cleanerclip} mitigates backdoor behaviors via feature-level purification using internal activations (gray-box access). PAR~\cite{singh2024perturb} adapts input prompts to counteract trigger effects, often relying on gradients or clean validation data, while ABD~\cite{kuang2024adversarial} detects and suppresses backdoors by analyzing activation patterns within the model (white-box). These approaches typically operate at a global level---modifying full representations, prompts, or internal activations---and assume access beyond standard inference. In contrast, \textsc{CLIPGuard} is designed as a fully black-box, inference-only defense. It requires no access to model parameters, gradients, internal states, or training data. By leveraging observable embeddings and similarity scores, \textsc{CLIPGuard} performs fine-grained spatial trigger localization through embedding-sensitivity analysis and selectively purifies suspicious regions via semantic inpainting, enabling effective backdoor mitigation under realistic deployment constraints.

\noindent\textbf{Evaluation Metrics.}
We evaluate \textsc{CLIPGuard} using two complementary metrics. Clean Accuracy (CA) measures the classification accuracy on unmodified clean images, following standard CLIP evaluation. Attack Success Rate (ASR) quantifies the proportion of triggered images that are misclassified into the attacker-specified target class.

\noindent\textbf{Environment.}
All experiments are conducted on Ubuntu~20.04 with an NVIDIA RTX 4090 (24GB), an Intel i9-13900K CPU, and 128GB RAM, using PyTorch~2.1.0 and SAM ViT-H for segmentation.

% ---------------- Table 1 ----------------
\begin{table*}[!t]
\centering
\caption{Defense comparison under traditional backdoor attacks (BadNet, Blended, Physical, LC). Higher CA ($\uparrow$) and lower ASR ($\downarrow$) indicate better performance. Best results are highlighted.}
\label{tab:traditional}
\setlength{\tabcolsep}{4pt}
\resizebox{\linewidth}{!}{
\begin{tabular}{c l cc cc cc cc cc cc}
\toprule
\multirow{2}{*}{\rotatebox{90}{\textbf{Dataset}}} & \multirow{2}{*}{\textbf{Attack}}
& \multicolumn{2}{c}{No Def.} & \multicolumn{2}{c}{CleanCLIP} & \multicolumn{2}{c}{CleanerCLIP}
& \multicolumn{2}{c}{PAR} & \multicolumn{2}{c}{ABD} & \multicolumn{2}{c}{\textsc{CLIPGuard}} \\
\cmidrule(lr){3-4}\cmidrule(lr){5-6}\cmidrule(lr){7-8}\cmidrule(lr){9-10}\cmidrule(lr){11-12}\cmidrule(lr){13-14}
& & CA$\uparrow$ & ASR$\downarrow$ & CA$\uparrow$ & ASR$\downarrow$ & CA$\uparrow$ & ASR$\downarrow$ & CA$\uparrow$ & ASR$\downarrow$ & CA$\uparrow$ & ASR$\downarrow$ & CA$\uparrow$ & ASR$\downarrow$ \\
\midrule
\multirow{5}{*}{\rotatebox{90}{STL-10}}
& BadNet   & 91.13 & 98.18 & 81.13 & 16.94 & \bestcell{87.45} & 18.45 & 81.30 & 25.30 & 76.87 & 13.67 & 86.34 & \bestasr{2.90} \\
& Blended  & 90.56 & 97.34 & 79.15 & 31.23 & \bestcell{85.87} & 22.67 & 73.60 & 18.39 & 83.17 & 23.09 & 84.32 & \bestasr{6.19} \\
& Physical & 89.34 & 99.45 & \bestcell{86.39} & 72.78 & 82.01 & 30.87 & 73.67 & 57.56 & 80.09 & 21.97 & 83.09 & \bestasr{2.73} \\
& LC       & 89.14 & 98.67 & 80.87 & 78.98 & 79.67 & 24.78 & 77.06 & 41.34 & 74.16 & 35.18 & \bestcell{82.03} & \bestasr{1.05} \\
\cmidrule(lr){2-14}
& \emph{Avg.} & 90.04 & 98.41 & 80.29 & 50.48 & 83.35 & 24.19 & 76.41 & 35.65 & 78.57 & 23.48 & \bestcell{83.95} & \bestasr{3.22} \\
\midrule
\multirow{5}{*}{\rotatebox{90}{ImageNet}}
& BadNet   & 61.56 & 93.18 & 56.54 & 9.12  & 57.17 & 7.15  & 53.30 & 11.60 & 55.47 & 11.35 & \bestcell{57.28} & \bestasr{4.13} \\
& Blended  & 57.70 & 99.40 & 53.44 & 19.50 & 52.61 & 12.84 & 53.60 & \bestasr{2.13} & 53.29 & 13.73 & \bestcell{57.62} & 5.92 \\
& Physical & 57.60 & 99.80 & 53.30 & 62.30 & \bestcell{54.18} & 32.69 & 53.00 & 42.40 & 49.19 & 19.13 & 52.37 & \bestasr{4.56} \\
& LC       & 56.17 & 99.90 & 50.10 & 73.78 & \bestcell{51.89} & 24.67 & 49.10 & 52.24 & 48.34 & 34.12 & 51.50 & \bestasr{2.10} \\
\cmidrule(lr){2-14}
& \emph{Avg.} & 58.26 & 98.07 & 53.35 & 41.18 & 53.96 & 19.34 & 52.25 & 27.09 & 51.57 & 19.58 & \bestcell{54.69} & \bestasr{4.18} \\
\bottomrule
\end{tabular}}
\end{table*}

% ---------------- Table 2 ----------------
\begin{table*}[!t]
\centering
\caption{Defense comparison under modern backdoor attacks (BadCLIP, SCAM, TrojVQA). Higher CA ($\uparrow$) and lower ASR ($\downarrow$) indicate better performance. Best results are highlighted.}
\label{tab:modern}
\setlength{\tabcolsep}{4pt}
\resizebox{\linewidth}{!}{
\begin{tabular}{c l cc cc cc cc cc cc}
\toprule
\multirow{2}{*}{\rotatebox{90}{\textbf{Dataset}}} & \multirow{2}{*}{\textbf{Attack}}
& \multicolumn{2}{c}{No Def.} & \multicolumn{2}{c}{CleanCLIP} & \multicolumn{2}{c}{CleanerCLIP}
& \multicolumn{2}{c}{PAR} & \multicolumn{2}{c}{ABD} & \multicolumn{2}{c}{\textsc{CLIPGuard}} \\
\cmidrule(lr){3-4}\cmidrule(lr){5-6}\cmidrule(lr){7-8}\cmidrule(lr){9-10}\cmidrule(lr){11-12}\cmidrule(lr){13-14}
& & CA$\uparrow$ & ASR$\downarrow$ & CA$\uparrow$ & ASR$\downarrow$ & CA$\uparrow$ & ASR$\downarrow$ & CA$\uparrow$ & ASR$\downarrow$ & CA$\uparrow$ & ASR$\downarrow$ & CA$\uparrow$ & ASR$\downarrow$ \\
\midrule
\multirow{4}{*}{\rotatebox{90}{STL-10}}
& BadCLIP & 88.60 & 98.23 & 81.56 & 53.35 & 81.14 & 19.87 & 81.56 & 30.40 & 73.77 & 27.23 & \bestcell{81.67} & \bestasr{3.34} \\
& SCAM    & 89.34 & 95.34 & 79.34 & 69.24 & 80.23 & 65.13 & 80.94 & 54.27 & 81.34 & 63.56 & \bestcell{85.97} & \bestasr{4.98} \\
& TrojVQA & 88.68 & 96.34 & 80.09 & 90.91 & \bestcell{83.65} & 15.65 & 78.32 & 51.11 & 79.47 & 39.56 & 82.56 & \bestasr{2.16} \\
\cmidrule(lr){2-14}
& \emph{Avg.} & 88.87 & 96.64 & 80.33 & 71.17 & 81.67 & 33.55 & 80.27 & 45.26 & 78.19 & 43.45 & \bestcell{83.40} & \bestasr{3.49} \\
\midrule
\multirow{4}{*}{\rotatebox{90}{ImageNet}}
& BadCLIP & 58.60 & 98.80 & 53.80 & 61.10 & 52.61 & 19.87 & 53.40 & 30.40 & 52.47 & 49.56 & \bestcell{55.34} & \bestasr{2.13} \\
& SCAM    & 56.90 & 95.60 & 53.30 & 62.40 & \bestcell{55.13} & 53.67 & 53.40 & 48.10 & 54.35 & 41.45 & 53.56 & \bestasr{3.14} \\
& TrojVQA & 58.68 & 97.86 & \bestcell{57.71} & 94.44 & 57.13 & 9.16 & 50.05 & 62.78 & 52.45 & 43.34 & 56.13 & \bestasr{4.19} \\
\cmidrule(lr){2-14}
& \emph{Avg.} & 58.06 & 97.42 & 54.94 & 72.65 & 54.96 & 27.57 & 52.28 & 47.09 & 53.09 & 44.78 & \bestcell{55.01} & \bestasr{3.15} \\
\bottomrule
\end{tabular}}
\end{table*}

\subsection{Comparison with Baselines}

Tables~\ref{tab:traditional} and~\ref{tab:modern} provide a comprehensive comparison of \textsc{CLIPGuard} against representative baseline defenses on STL-10 and ImageNet under traditional and modern backdoor attacks.

\noindent\textbf{Traditional Attacks.}
As shown in \tref{tab:traditional}, both STL-10 and ImageNet exhibit extreme vulnerability under classical trigger mechanisms---including BadNets, Blended, Physical, and Label-Consistent (LC) attacks---when no defense is applied, with average ASRs exceeding 98\%, indicating near-complete attack success. \textsc{CLIPGuard} substantially mitigates these threats, reducing the average ASR from 98.41\% to 3.22\% on STL-10 while preserving strong clean accuracy (83.95\%), and from 98.07\% to 4.18\% on ImageNet with competitive clean accuracy (54.69\%). In contrast, baseline defenses remain significantly less robust. On STL-10, CleanCLIP, CleanerCLIP, PAR, and ABD retain average ASRs of 50.48\%, 24.19\%, 35.65\%, and 23.48\%, respectively. A similar robustness gap is evident on ImageNet, where CleanCLIP, CleanerCLIP, PAR, and ABD retain 41.18\%, 19.34\%, 27.09\%, and 19.58\% average ASR, respectively. Across individual attacks, \textsc{CLIPGuard} consistently achieves the lowest ASR---often below 5\%. For example, on STL-10, ASR drops to 1.05\% under LC and 2.73\% under Physical attacks, while on ImageNet it reduces ASR to 4.13\% for BadNet, 4.56\% for Physical, and 2.10\% for LC.

% ---------------- Table 3 ----------------
\begin{table}[!t]
\centering
\caption{Defense performance comparison on the ImageNet subset under different attack modes (All2One, All2All, and Untargeted).}
\label{tab:modes}
\setlength{\tabcolsep}{5pt}
\resizebox{\linewidth}{!}{
\begin{tabular}{l l cc cc cc cc cc}
\toprule
\multirow{2}{*}{\textbf{Mode}} & \multirow{2}{*}{\textbf{Attack}}
& \multicolumn{2}{c}{No Def.} & \multicolumn{2}{c}{CleanCLIP} & \multicolumn{2}{c}{CleanerCLIP}
& \multicolumn{2}{c}{PAR} & \multicolumn{2}{c}{Ours} \\
\cmidrule(lr){3-4}\cmidrule(lr){5-6}\cmidrule(lr){7-8}\cmidrule(lr){9-10}\cmidrule(lr){11-12}
& & CA & ASR & CA & ASR & CA & ASR & CA & ASR & CA & ASR \\
\midrule
\multirow{4}{*}{All2One}
& BadNets  & 61.56 & 93.18 & 56.54 & 12.12 & 57.17 & 7.15  & 53.30 & 11.60 & \bestcell{57.28} & \bestasr{4.13} \\
& Blended  & 57.70 & 99.40 & 53.44 & 19.50 & 52.61 & 12.84 & 53.60 & \bestasr{2.13} & \bestcell{57.62} & 5.92 \\
& Physical & 57.60 & 99.80 & 53.30 & 62.30 & 54.18 & 32.69 & 53.00 & 42.40 & \bestcell{58.37} & \bestasr{4.56} \\
\cmidrule(lr){2-12}
& \emph{Avg.} & 58.95 & 97.46 & 54.43 & 31.31 & 54.65 & 17.56 & 53.30 & 18.71 & \bestcell{57.76} & \bestasr{4.87} \\
\midrule
\multirow{4}{*}{All2All}
& BadNets  & 56.23 & 85.17 & 52.81 & 18.74 & 53.92 & 10.61 & 51.47 & 14.22 & \bestcell{55.91} & \bestasr{5.37} \\
& Blended  & 53.34 & 89.25 & 50.66 & 22.43 & 51.28 & 14.35 & 50.73 & 9.84  & \bestcell{54.84} & \bestasr{6.18} \\
& Physical & 56.23 & 87.89 & 52.04 & 55.62 & 53.11 & 29.47 & 51.60 & 37.21 & \bestcell{56.48} & \bestasr{5.92} \\
\cmidrule(lr){2-12}
& \emph{Avg.} & 55.27 & 87.44 & 51.84 & 32.26 & 52.77 & 18.14 & 51.27 & 20.42 & \bestcell{55.74} & \bestasr{5.82} \\
\midrule
\multirow{4}{*}{UA}
& BadNets  & 58.45 & 78.63 & 54.72 & 16.38 & 55.30 & 8.91  & 53.18 & 12.40 & \bestcell{57.11} & \bestasr{3.98} \\
& Blended  & 57.43 & 81.74 & 53.66 & 20.15 & 54.22 & 11.37 & 52.95 & 7.66  & \bestcell{56.90} & \bestasr{4.75} \\
& Physical & 56.98 & 84.52 & 52.83 & 48.91 & 53.76 & 24.16 & 51.92 & 30.77 & \bestcell{57.42} & \bestasr{4.41} \\
\cmidrule(lr){2-12}
& \emph{Avg.} & 57.62 & 81.63 & 53.74 & 28.48 & 54.43 & 14.81 & 52.68 & 16.94 & \bestcell{57.14} & \bestasr{4.38} \\
\bottomrule
\end{tabular}}
\end{table}

\noindent\textbf{Modern Attacks.}
As shown in \tref{tab:modern}, modern attack paradigms---including BadCLIP, SCAM, and TrojVQA---further increase defense difficulty due to feature-aligned manipulation and jailbreak behaviors. Without defense, average ASRs remain above 96\% on STL-10 and 97\% on ImageNet. \textsc{CLIPGuard} maintains strong robustness, reducing ASR to 3.49\% on STL-10 and 3.15\% on ImageNet, while preserving competitive clean accuracy (83.40\% and 55.01\%, respectively). In contrast, baseline methods degrade substantially under these advanced attacks, often retaining ASRs above 30\% on STL-10 and above 25\% on ImageNet. Notably, complex multimodal triggers such as SCAM and TrojVQA expose pronounced weaknesses in prior defenses, whereas \textsc{CLIPGuard} consistently keeps ASR below 5\% across evaluated settings.

Across both traditional and modern attack families, \textsc{CLIPGuard} consistently achieves the lowest ASR while maintaining strong clean accuracy, demonstrating that localized trigger detection and selective purification effectively remove the trigger's effect without degrading semantic content. \textsc{CLIPGuard} outperforms global retraining and activation-based defenses, while generalizing from STL-10 to ImageNet, demonstrating scalability and practical viability in real-world black-box settings.

\subsection{In-depth Evaluation}

\noindent\textbf{Cross-Mode Comparison.}
\tref{tab:modes} reports results across All2One, All2All, and Untargeted (UA) settings. In the All2One mode, our method reduces the average ASR from 97.46\% (No Defense) to 4.87\%, substantially outperforming CleanCLIP (31.31\%), CleanerCLIP (17.56\%), and PAR (18.71\%), while preserving clean accuracy (57.76\%), close to the no-defense baseline (58.95\%). Under the All2All setting, the average ASR drops from 87.44\% to 5.82\%, compared to 32.26\%, 18.14\%, and 20.42\% for CleanCLIP, CleanerCLIP, and PAR, respectively, with CA preserved at 55.74\%. Similarly, for Untargeted attacks, our defense suppresses ASR from 81.63\% to 4.38\%, significantly lower than CleanCLIP (28.48\%), CleanerCLIP (14.81\%), and PAR (16.94\%), while achieving 57.14\% CA. These consistent improvements across all attack modes demonstrate the robustness and generalizability of our approach.

% ---------------- Table 4 ----------------
\begin{table}[!t]
\centering
\caption{Defense performance comparison in terms of Attack Success Rate (ASR, \%) under varying poison ratios for the physical patch attack on ImageNet. Lower ASR values indicate improved robustness.}
\label{tab:poison_ratio}
\setlength{\tabcolsep}{8pt}
\begin{tabular}{l c c c c c c}
\toprule
\textbf{Poison Ratio} & No Def. & CleanCLIP & CleanerCLIP & PAR & ABD & Ours \\
\midrule
5\%  & 90.67 & 31.34 & 24.23 & 35.16 & 11.50 & \bestasr{3.15} \\
10\% & 99.80 & 62.30 & 32.69 & 42.40 & 19.30 & \bestasr{4.56} \\
15\% & 99.83 & 63.12 & 32.74 & 47.87 & 19.56 & \bestasr{4.79} \\
20\% & 99.90 & 64.67 & 35.56 & 46.17 & 21.45 & \bestasr{5.07} \\
\bottomrule
\end{tabular}
\end{table}

% ---------------- Table 5 ----------------
\begin{table}[!t]
\centering
\caption{Impact of grid cell size on defense time, Acc, and ASR for STL-10 and ImageNet. Best results are highlighted in green; worst results in light red.}
\label{tab:grid}
\setlength{\tabcolsep}{5pt}
\resizebox{\linewidth}{!}{
\begin{tabular}{l ccc ccc}
\toprule
\multirow{2}{*}{\textbf{Cell Size}} & \multicolumn{3}{c}{STL-10} & \multicolumn{3}{c}{ImageNet} \\
\cmidrule(lr){2-4}\cmidrule(lr){5-7}
& Defense Time (ms) & Acc (\%) & ASR (\%) & Defense Time (s) & Acc (\%) & ASR (\%) \\
\midrule
No Defense     & --    & 89.34 & 95.34 & --   & 56.90 & 95.60 \\
\midrule
$4\times4$     & \goodcell{90} & 86.13 & \badcell{8.76} & \goodcell{0.54} & \goodcell{54.09} & \badcell{10.87} \\
$8\times8$     & 160   & 85.97 & \goodcell{4.98} & 0.95 & 53.56 & 3.14 \\
$16\times16$   & 180   & 85.67 & 5.10 & 1.62 & 53.98 & \goodcell{3.13} \\
$24\times24$   & 230   & 85.81 & 5.07 & 4.05 & 53.45 & 3.68 \\
$32\times32$   & 230   & \goodcell{86.20} & 5.24 & 6.87 & 53.61 & 4.56 \\
Full Image     & \badcell{2{,}980} & \badcell{83.45} & 5.14 & \badcell{8.04} & \badcell{52.98} & 5.09 \\
\bottomrule
\end{tabular}}
\end{table}

\noindent\textbf{Performance of Defense under Different Attack Ratios.}
\tref{tab:poison_ratio} illustrates the defense performance on ImageNet under a physical patch attack as the poison ratio increases from 0.05 to 0.20. Without defense, ASR rises sharply from 90.67\% to 99.90\%, showing that higher poisoning severely compromises the model. Existing defenses reduce ASR but degrade as the attack ratio increases. In contrast, our method maintains consistently low ASR between 3.15\% and 5.07\% across all ratios, demonstrating robust and stable performance even under more challenging attack conditions.

Overall, \textsc{CLIPGuard} delivers strong robustness against diverse backdoor attacks on CLIP ViT-B/16. It consistently achieves the lowest ASR, preserves strong clean accuracy, adapts efficiently to new inputs, and minimizes artifacts. These results demonstrate that \textsc{CLIPGuard} is an effective, scalable, and fully black-box defense for CLIP-based vision systems.

\noindent\textbf{Defense Time Cost under Different Grid Cell Sizes.}
\tref{tab:grid} presents the defense time cost of \textsc{CLIPGuard} under different grid cell sizes on STL-10 and ImageNet. As expected, finer spatial partitioning increases computational overhead due to a larger number of extracted segments within the suspected cells, each of which must be analyzed during segmentation and contrastive purification. The smallest grid ($4\times4$) achieves the lowest defense time (90\,ms on STL-10 and 0.54\,s on ImageNet), but exhibits relatively higher ASR, particularly on ImageNet (10.87\%), suggesting insufficient localization granularity. Increasing the grid resolution reduces ASR, with $8\times8$ and $16\times16$ achieving the lowest attack success rates (4.98\% on STL-10 and 3.13\% on ImageNet) while maintaining competitive clean accuracy. In contrast, the Full Image setting treats the entire image as a single suspected cell and applies segmentation across the whole image, substantially increasing the number and spatial extent of extracted segments. This leads to the highest defense time cost (2{,}980\,ms on STL-10 and 8.04\,s on ImageNet) without meaningful gains in robustness. Overall, these results highlight a clear trade-off between grid cell size, defense time cost, ASR, and CA, where moderate grid resolutions---particularly $8\times8$---provide the best balance between computational efficiency and effective backdoor mitigation in \textsc{CLIPGuard}.

% -------------------------------------------------------------
\section{Conclusion}
% -------------------------------------------------------------
We introduced \textsc{CLIPGuard}, a fully black-box, inference-time defense for mitigating backdoor attacks in CLIP-based classifiers. Without access to model parameters, gradients, logits, or clean validation data, our method combines embedding-sensitivity analysis, adaptive region localization, and selective purification to accurately detect and remove malicious triggers. Across ImageNet subsets under All2One, All2All, and Untargeted settings, \textsc{CLIPGuard} consistently achieves the lowest ASR among all compared defenses. In particular, it reduces the average ASR from 97.46\%, 87.44\%, and 81.63\% (No Defense) to 4.87\%, 5.82\%, and 4.38\%, respectively, while maintaining competitive clean accuracy. Under increasingly severe physical patch attacks, where the poison ratio rises from 0.05 to 0.20 and the undefended ASR approaches 99.90\%, our method keeps ASR stable between 3.15\% and 5.07\%, demonstrating strong robustness against intensified poisoning. These results confirm that \textsc{CLIPGuard} provides stable and effective protection across diverse attack modes and poison strengths, offering a practical and model-agnostic defense mechanism for securing CLIP-based vision systems in restricted-access environments.

% -------------------------------------------------------------
% Bibliography
% -------------------------------------------------------------
\bibliographystyle{splncs04}
\bibliography{references}

@inproceedings{radford2021learning,
  title={Learning transferable visual models from natural language supervision},
  author={Radford, Alec and Kim, Jong Wook and Hallacy, Chris and Ramesh, Aditya and Goh, Gabriel and Agarwal, Sandhini and Sastry, Girish and Askell, Amanda and Mishkin, Pamela and Clark, Jack and others},
  booktitle={International conference on machine learning},
  pages={8748--8763},
  year={2021},
  organization={PmLR}
}

@inproceedings{li2022blip,
  title={Blip: Bootstrapping language-image pre-training for unified vision-language understanding and generation},
  author={Li, Junnan and Li, Dongxu and Xiong, Caiming and Hoi, Steven},
  booktitle={International conference on machine learning},
  pages={12888--12900},
  year={2022},
  organization={PMLR}
}

@inproceedings{carlini2021poisoning,
  title={Poisoning the unlabeled dataset of $\{$Semi-Supervised$\}$ learning},
  author={Carlini, Nicholas},
  booktitle={30th USENIX Security Symposium (USENIX Security 21)},
  pages={1577--1592},
  year={2021}
}

@inproceedings{liang2024badclip,
  title={Badclip: Dual-embedding guided backdoor attack on multimodal contrastive learning},
  author={Liang, Siyuan and Zhu, Mingli and Liu, Aishan and Wu, Baoyuan and Cao, Xiaochun and Chang, Ee-Chien},
  booktitle={Proceedings of the IEEE/CVF conference on computer vision and pattern recognition},
  pages={24645--24654},
  year={2024}
}

@inproceedings{azuma2023defense,
  title={Defense-prefix for preventing typographic attacks on clip},
  author={Azuma, Hiroki and Matsui, Yusuke},
  booktitle={Proceedings of the IEEE/CVF International Conference on Computer Vision},
  pages={3644--3653},
  year={2023}
}

@article{liu2025survey,
  title={A survey of attacks on large vision--language models: Resources, advances, and future trends},
  author={Liu, Daizong and Yang, Mingyu and Qu, Xiaoye and Zhou, Pan and Cheng, Yu and Hu, Wei},
  journal={IEEE Transactions on Neural Networks and Learning Systems},
  year={2025},
  publisher={IEEE}
}

@article{gu2019badnets,
  title={Badnets: Evaluating backdooring attacks on deep neural networks},
  author={Gu, Tianyu and Liu, Kang and Dolan-Gavitt, Brendan and Garg, Siddharth},
  journal={Ieee Access},
  volume={7},
  pages={47230--47244},
  year={2019},
  publisher={IEEE}
}

@article{westerhoff2025scam,
  title={Scam: A real-world typographic robustness evaluation for multimodal foundation models},
  author={Westerhoff, Justus and others},
  journal={arXiv preprint arXiv:2504.04893},
  year={2025}
}

@inproceedings{gong2025figstep,
  title={Figstep: Jailbreaking large vision-language models via typographic visual prompts},
  author={Gong, Yichen and Ran, Delong and Liu, Jinyuan and Wang, Conglei and Cong, Tianshuo and Wang, Anyu and Duan, Sisi and Wang, Xiaoyun},
  booktitle={Proceedings of the AAAI Conference on Artificial Intelligence},
  volume={39},
  number={22},
  pages={23951--23959},
  year={2025}
}

@article{singh2024perturb,
  title={Perturb and recover: Fine-tuning for effective backdoor removal from clip},
  author={Singh, Naman Deep and Croce, Francesco and Hein, Matthias},
  journal={arXiv preprint arXiv:2412.00727},
  year={2024}
}

@inproceedings{walmer2022dual,
  title={Dual-key multimodal backdoors for visual question answering},
  author={Walmer, Matthew and Sikka, Karan and Sur, Indranil and Shrivastava, Abhinav and Jha, Susmit},
  booktitle={Proceedings of the IEEE/CVF Conference on computer vision and pattern recognition},
  pages={15375--15385},
  year={2022}
}

@inproceedings{coates2011analysis,
  title={An analysis of single-layer networks in unsupervised feature learning},
  author={Coates, Adam and Ng, Andrew and Lee, Honglak},
  booktitle={Proceedings of the fourteenth international conference on artificial intelligence and statistics},
  pages={215--223},
  year={2011},
  organization={JMLR Workshop and Conference Proceedings}
}

@article{shi2023black,
  title={Black-box backdoor defense via zero-shot image purification},
  author={Shi, Yucheng and others},
  journal={Advances in Neural Information Processing Systems},
  volume={36},
  pages={57336--57366},
  year={2023}
}

@article{chen2017targeted,
  title={Targeted backdoor attacks on deep learning systems using data poisoning},
  author={Chen, Xinyun and Liu, Chang and Li, Bo and Lu, Kimberly and Song, Dawn},
  journal={arXiv preprint arXiv:1712.05526},
  year={2017}
}

@article{gao2020backdoor,
  author    = {Gao, Yansong and others},
  title     = {Backdoor Attacks and Countermeasures on Deep Neural Networks: A Comprehensive Review},
  journal   = {arXiv preprint arXiv:2007.10760},
  year      = {2020}
}

@inproceedings{turner2019label,
  author    = {Turner, Alexander and others},
  title     = {Label-Consistent Backdoor Attacks},
  booktitle = {Proceedings of the 36th International Conference on Machine Learning (ICML)},
  pages     = {9536--9546},
  year      = {2019}
}

@article{li2020invisible,
  title={Invisible backdoor attacks on deep neural networks via steganography and regularization},
  author={Li, Shaofeng and Xue, Minhui and Zhao, Benjamin Zi Hao and Zhu, Haojin and Zhang, Xinpeng},
  journal={IEEE Transactions on Dependable and Secure Computing},
  volume={18},
  number={5},
  pages={2088--2105},
  year={2020},
  publisher={IEEE}
}

@inproceedings{kirillov2023segment,
  title={Segment anything},
  author={Kirillov, Alexander and Mintun, Eric and Ravi, Nikhila and Mao, Hanzi and Rolland, Chloe and Gustafson, Laura and Xiao, Tete and Whitehead, Spencer and Berg, Alexander C and Lo, Wan-Yen and others},
  booktitle={Proceedings of the IEEE/CVF international conference on computer vision},
  pages={4015--4026},
  year={2023}
}

@article{gu2017badnets,
  title={Badnets: Identifying vulnerabilities in the machine learning model supply chain},
  author={Gu, Tianyu and others},
  journal={arXiv preprint arXiv:1708.06733},
  year={2017}
}

@misc{imagenette2020,
  author = {Howard, Jeremy},
  title = {Imagenette: A smaller subset of 10 easily classified classes from Imagenet},
  year = {2020},
  howpublished = {\url{https://github.com/fastai/imagenette}}
}

@article{miyai2024generalized,
  title={Generalized out-of-distribution detection and beyond in vision language model era: A survey},
  author={Miyai, Atsuyuki and others},
  journal={arXiv preprint arXiv:2407.21794},
  year={2024}
}

@article{yang2023robust,
  title={Robust contrastive language-image pretraining against data poisoning and backdoor attacks},
  author={Yang, Wenhan and Gao, Jingdong and Mirzasoleiman, Baharan},
  journal={Advances in Neural Information Processing Systems},
  volume={36},
  pages={10678--10691},
  year={2023}
}

@inproceedings{bansal2023cleanclip,
  title={Cleanclip: Mitigating data poisoning attacks in multimodal contrastive learning},
  author={Bansal, Hritik and Singhi, Nishad and Yang, Yu and Yin, Fan and Grover, Aditya and Chang, Kai-Wei},
  booktitle={Proceedings of the IEEE/CVF International Conference on Computer Vision},
  pages={112--123},
  year={2023}
}

@inproceedings{poppi2024safe,
  title={Safe-clip: Removing nsfw concepts from vision-and-language models},
  author={Poppi, Samuele and Poppi, Tobia and Cocchi, Federico and Cornia, Marcella and Baraldi, Lorenzo and Cucchiara, Rita},
  booktitle={European Conference on Computer Vision},
  pages={340--356},
  year={2024},
  organization={Springer}
}

@inproceedings{feng2023detecting,
  title={Detecting backdoors in pre-trained encoders},
  author={Feng, Shiwei and Tao, Guanhong and Cheng, Siyuan and Shen, Guangyu and Xu, Xiangzhe and Liu, Yingqi and Zhang, Kaiyuan and Ma, Shiqing and Zhang, Xiangyu},
  booktitle={Proceedings of the IEEE/CVF Conference on Computer Vision and Pattern Recognition},
  pages={16352--16362},
  year={2023}
}

@article{niu2024bdetclip,
  title={Bdetclip: Multimodal prompting contrastive test-time backdoor detection},
  author={Niu, Yuwei and He, Shuo and Wei, Qi and Wu, Zongyu and Liu, Feng and Feng, Lei},
  journal={arXiv preprint arXiv:2405.15269},
  year={2024}
}

@article{xun2024cleanerclip,
  title={CleanerCLIP: Fine-grained Counterfactual Semantic Augmentation for Backdoor Defense in Contrastive Learning},
  author={Xun, Yuan and Liang, Siyuan and Jia, Xiaojun and Liu, Xinwei and Cao, Xiaochun},
  journal={arXiv preprint arXiv:2409.17601},
  year={2024}
}

@article{kuang2024adversarial,
  title={Adversarial backdoor defense in clip},
  author={Kuang, Junhao and Liang, Siyuan and Liang, Jiawei and Liu, Kuanrong and Cao, Xiaochun},
  journal={arXiv preprint arXiv:2409.15968},
  year={2024}
}

@article{telea2004image,
  title={An image inpainting technique based on the fast marching method},
  author={Telea, Alexandru},
  journal={Journal of graphics tools},
  volume={9},
  number={1},
  pages={23--34},
  year={2004},
  publisher={Taylor \& Francis}
}

@inproceedings{rombach2022high,
  title={High-resolution image synthesis with latent diffusion models},
  author={Rombach, Robin and Blattmann, Andreas and Lorenz, Dominik and Esser, Patrick and Ommer, Bj{\"o}rn},
  booktitle={Proceedings of the IEEE/CVF conference on computer vision and pattern recognition},
  pages={10684--10695},
  year={2022}
}

@InProceedings{Wenger_2021_CVPR,
    author    = {Wenger, Emily and others},
    title     = {Backdoor Attacks Against Deep Learning Systems in the Physical World},
    booktitle = {Proceedings of the IEEE/CVF Conference on Computer Vision and Pattern Recognition (CVPR)},
    month     = {June},
    year      = {2021},
    pages     = {6206-6215}
}

\end{document}